# SKIN-COLOR BASED VIDEOS CATEGORIZATION

Rehanullah Khan [1], Asad Maqsood [1], Zeeshan Khan [2], Muhammad Ishaq [3], Arsalan Arif [1]

[1] Sarhad University of Science and Information Technology, Peshawar, Pakistan
[2] RWTH Aachen, The Chair of Human Language Technology and Pattern Recognition
[3] UET Mardan, Peshawar, Pakistan

ABSTRACT

On dedicated websites, people can upload videos and share it with the rest of the world. Currently these videos are categorized manually by the help of the user community. In this paper, we propose a combination of color spaces with the Bayesian network approach for robust detection of skin color followed by an automated video categorization. Experimental results show that our method can achieve satisfactory performance for categorizing videos based on skin color.

## 1. INTRODUCTION

Locating and tracking patches of skin-colored pixels through an image is a tool used in many face recognition and gesture tracking systems [13][8]. Skin information contributes much to object recognition [18]. One of the usage of skin color based tracking, locating and categorization could be blocking unwanted video contents on World Wide Web. On dedicated websites, people can upload videos and share it with the rest of the world. There are uploaded adult videos, which may not be allowed by the service providers. Therefore, how to effectively categorize and block such videos has been arousing a serious concern for the service providers.

The mostly used approach to contents blocking on the Internet is based on contextual keyword pattern matching technology that categorizes URLs by means of checking contexts of web pages or video names and then traps the websites [11][15]. This does not hold true for websites which allow uploading videos like Google Videos and YouTube, because the videos uploaded have different names from the contents they contain. Due to no automated process, the Google and YouTube rely on user's community. Therefore, an automated method to detect and categorize videos based on skin color will help the service providers and can provide control over the videos contents.

According to Smeulders et. al [14] color has been an active area of research in image retrieval, more than in any other branch of computer vision. The interest in color may be ascribed to the superior discriminating potentiality of a three dimensional domain compared to the single dimensional domain of gray-level images [14].

The goal of our system is to categorize videos based on skin color. Depending on the percentage of skin in videos, the videos are flagged as Large-Skin-Videos (LSKIN), Partial-Skin-Videos (PSKIN) and No-Skin-Videos (NSKIN). The set of videos used in our experiments consists of 30 videos, collected and provided by video service provider. The service provider defines successful categorization as a true positive rate of above 70% because this would decrease the amount of manual work dramatically.

The remainder of the paper is organized as follows: Section 2 explains the previous work. Section 3 discusses color spaces used and the algorithm. Section 4 discusses experimental results and Section 5 concludes this paper.

## 2. PREVIOUS WORK

Singh et al. [13] discusses in detail different color spaces and skin detection. In their work, three color spaces; RGB, YCbCr and HSI are of main concern. They have compared the algorithms based on these color spaces and have combined them for face detection. The algorithm fails when sufficient non face skin is visible in the images. In [16], color spaces and their output results for skin detection are discussed. Furthermore, they state that excluding color luminance from the classification process cannot help achieving better discrimination of skin and non skin.

In [19] and [20], image filters based on skin information are described. The first step in their approach is skin detection. Maximum entropy modeling is used to model the distribution of skinness from the input image. A first order model is built that introduces constraints on color gradients of neighboring pixels. The output of skin detection is a gray scale skin map with the gray levels being proportional to the skin probabilities. There are false alarms when the background color matches human skin color. According to [3], a single color space may limit the performance of the skin color filter and that better performance can be achieved using two or more color spaces.

Jae et al. [10] discusses elliptical boundary model for skin color detection. To devise the appropriate model for skin detection, they investigate the characteristics of skin and non-skin distributions in terms of chrominance histograms. They don't take the advantage of combining different color spaces. In [17], a method to detect body parts in images is presented. The algorithm is composed of content-based and image-based classification approaches. In the content-based approach, color filtering and texture analysis is used to detect the skin region in an image and its classification depends on the presence of large skin bulks. In the image-based approach, the color histogram and coherence vectors are extracted to represent the color and spatial information of the image.

According to [11] and [12], the selection of color space influences the quality of skin color modeling. The pixels belonging to skin region exhibit similar Cb and Cr chromatic characteristics, therefore, the skin color model based on Cb and Cr values can provide good coverage of all human races. Accordingly, despite their different appearances, these color types belong to the same small cluster in Cb-Cr plane. The apparent difference in skin colors perceived by viewers mainly comes from the darkness or fairness of the skin. These features are reflected on the difference in the brightness of the color, which is governed by Y component rather than Cb and Cr components. It provides an effective separation into luminance and chrominance channel and generates a compact skin chroma distribution. Yang et al. [18] have introduced a new Gamma Correction method to weaken the effects of illumination on images and a new RGB nonlinear transformation to describe the skin and non-skin distributions. Khan et al. [9][4][6] use face detection for adapting to the changing illumination circumstances for skin detection in videos. The authors in [5] introduce the usage of Decision Trees for pixel based skin detection and classification. Skin detection based on global seeds is introduced in [7].

## 3. SKIN COLOR MODELING

Color is a low level feature, which makes it computationally inexpensive and therefore suitable for real-time object characterization, detection and localization [11]. The main goal of skin color detection or classification for skin contents filtering is to build a decision rule that will discriminate between skin and non-skin pixels. Identifying skin colored pixels involves finding the range of values for which most skin pixels would fall in a given color space. This may be as simple as explicitly classifying a pixel as a skin pixel if Red, Green and Blue color channels have specific value range distribution. Other techniques use Neural Networks and Bayesian methods [11].

### 3.1. RGB Color Space

In the RGB color space, each color appears in its primary spectral component of Red, Green and Blue. Images represented in the RGB space consist of three component images, one for each primary color. When fed into an RGB monitor, these images combine on the phosphor screen to produce a composite color image. The RGB color space is one of the most widely used color spaces for storing and processing digital image [16]. However, the RGB color space alone is not reliable for identifying skin colored pixels since it represents not only color but also luminance [16]. Skin luminance may vary within and across persons due to ambient lighting, therefore, it is not suitable for segmenting skin and non-skin regions. Chromatic colors are more reliable and are obtained by eliminating luminance through nonlinear transformations [16].

### 3.2. YCbCr

YCbCr is an encoded nonlinear RGB signal, commonly used by European Television Studios and for image compression work [16]. Color is represented by luma which is luminance and computed from nonlinear RGB constructed as a weighted sum of the RGB values and two color difference values Cb and Cr that are formed by subtracting luma from RGB Red and Blue components. The transformation simplicity and explicit separation of luminance and chrominance components make this color space attractive for skin color modeling [16].

### 3.3. Skin Detection

For skin-color modeling, we construct a Bayesian network in the YCbCr and RGB color spaces. A Bayesian network is constructed from pixel triplet of the training skin colors.

A Bayesian network is also called a belief network and a directed acyclic graphical model. It is a representation for random variables and conditional independences within these random variables. The conditional independences are represented by Directed Acyclic Graph (DAG). More formally, a Bayesian network $B = <N, A, \theta>$ for skin color pixel (triplet) is a DAG $<N, A>$ with a conditional probability distribution for every node (collectively $\theta$ for all nodes). A node n N in the graph G represents some random variable, and each edge or each arc a A between nodes shows a probabilistic dependency. For learning Bayesian networks from specific datasets, data attributes are represented by nodes [1].

In a Bayesian network, the learner does not distinguish the skin and non-skin class variables from the attribute variables in data. As such, a network (or a set of networks) are created for skin color pixels that "best describes" the probability distribution of the training data. The problem of learning a Bayesian network can be stated as: Given a training set $D = \{u_1, ..., u_N\}$ of instances of U, find a network B that best matches D. Heuristic search techniques are used to find the best candidate in the space of possible networks. The search process relies on a scoring function that assesses the merits of each candidate network [2]. If we assume that for training, a Bayesian network B encodes a distribution $P_B(A_1, ..., A_n)$ from the training dataset with C classes, then for testing, a classifier based on B returns the label c that maximizes the posterior probability $P_B(c|a_1, ..., a_n)$. The network B can also be used to find out updated knowledge of the state of a subset of variables when other variables (the evidence variables) are observed.

### 3.4. Color-Space Intersection

The proposed skin categorization system starts with skin detection in videos based on the RGB color space. The detected skin pixels are passed to the YCbCr Bayesian detector. If the YCbCr skin detector confirms the pixels as skin pixels; the pixels are flagged as skin pixels. Depending on scenario and skin detected per frame, the video is flagged as LSKIN, PSKIN and NSKIN. Based on experiments, we set three rules for videos categorization: If the percentage of skin is greater than 15%, the video is flagged as LSKIN. If the skin percentage is greater than 3% and less than 15%, the video is flagged as PSKIN and NSKIN if less than 3%.

## 4. RESULTS

To evaluate the skin detection algorithm, we use a set of 30 challenging videos. Figure 1 shows example frames from these video sequences. The sequences span a wide range of environmental conditions. People of different ethnicity and various skin tones are represented. Sequences also contain scenes with multiple people and/or multiple visible body parts and scenes shot both indoors and outdoors, with moving camera. The lighting varies from natural light to directional stage lighting. Sequences contain shadows and minor occlusions. Videos in which background color matches the skin color are also present in the test set. Collected sequences vary in length from 100 frames to 1300 frames. These videos are divided into three categories depending on the amount of skin in video and serve as ground truth for the algorithm. Eleven videos are labeled as LSKIN, nine videos are labeled as PSKIN and ten 10 are labeled as NSKIN.

On the testing set, the algorithm correctly identified 28 out of 30 videos as shown in table 1. Figure 2(a) shows an example skin detection on a single frame from Video 1. Figure 2(a) shows skin detection in the YCbCr color space. Figure 2(b) indicates the peaks related to the correct identification of skin in the entire Video 1. Figure 3 shows an example frame and skin detection in the RGB color space. This example frame is extracted from Video 2 which is correctly categorized as LSKIN based on skin presence.

The algorithm incorrectly reported two videos, Video 12 and Video 24 as LSKIN. The reason being the skin colored objects (false skin colors) present in these videos. Figure 4 is an example frame from Video 12 which is categorized as

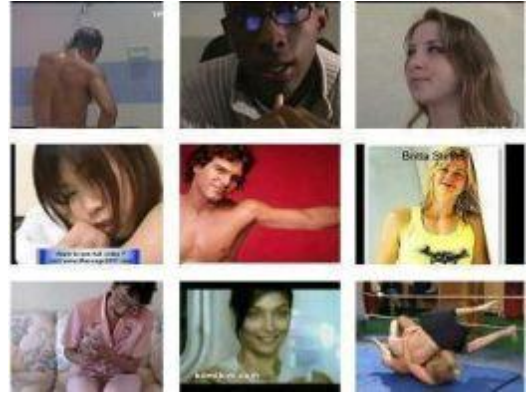

Fig. 1. Examples frames from video sequences used for experimentation.

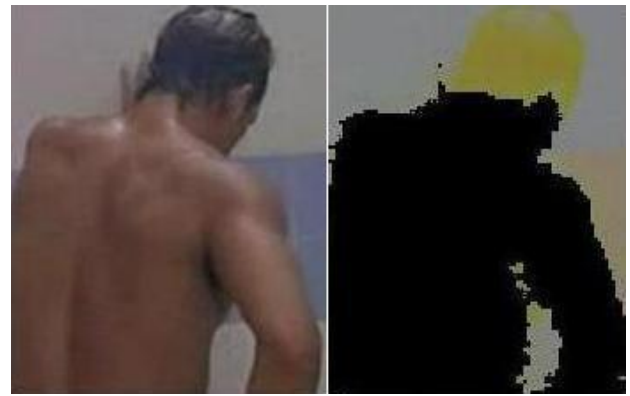

(a) Skin detection in the YCbCr color space (Video 1). Black shows skin.

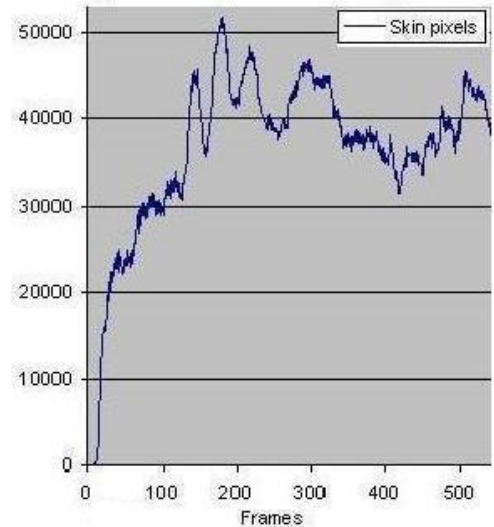

(b) Skin graph showing peaks for Video 1 regarding correct identification of skin.

Fig. 2. Skin detection scenario.

Table 1. Result of videos categorization. Bold indicates wrong classification by the algorithm.

| Video# | Ground Truth | Skin % | Result |
|---|---|---|---|
| 1 | LSKIN | 18.20 | LSKIN |
| 2 | LSKIN | 16.05 | LSKIN |
| 3 | LSKIN | 15.90 | LSKIN |
| 4 | LSKIN | 30.05 | LSKIN |
| 5 | LSKIN | 20.50 | LSKIN |
| 6 | LSKIN | 19.76 | LSKIN |
| 7 | LSKIN | 21.61 | LSKIN |
| 8 | LSKIN | 26.10 | LSKIN |
| 9 | LSKIN | 17.03 | LSKIN |
| 10 | LSKIN | 20.30 | LSKIN |
| 11 | LSKIN | 25.80 | LSKIN |
| **12** | **PSKIN** | **16.59** | **LSKIN** |
| 13 | PSKIN | 08.14 | PSKIN |
| 14 | PSKIN | 07.65 | PSKIN |
| 15 | PSKIN | 06.90 | PSKIN |
| 16 | PSKIN | 11.15 | PSKIN |
| 17 | PSKIN | 10.87 | PSKIN |
| 18 | PSKIN | 09.91 | PSKIN |
| 19 | PSKIN | 08.79 | PSKIN |
| 20 | PSKIN | 07.69 | PSKIN |
| 21 | NSKIN | 01.15 | NSKIN |
| 22 | NSKIN | 00.91 | NSKIN |
| 23 | NSKIN | 02.10 | NSKIN |
| **24** | **NSKIN** | **19.12** | **LSKIN** |
| 25 | NSKIN | 01.17 | NSKIN |
| 26 | NSKIN | 02.01 | NSKIN |
| 27 | NSKIN | 01.08 | NSKIN |
| 28 | NSKIN | 01.00 | NSKIN |
| 29 | NSKIN | 01.22 | NSKIN |
| 30 | NSKIN | 02.50 | NSKIN |

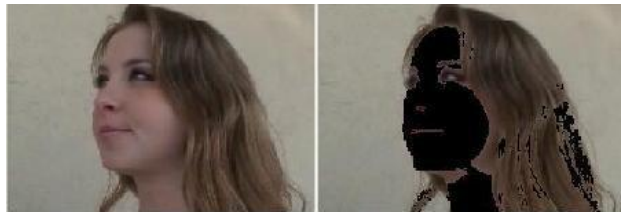

Fig. 3. Skin detection in the RGB color space on a single frame from Video 2. Black shows skin.

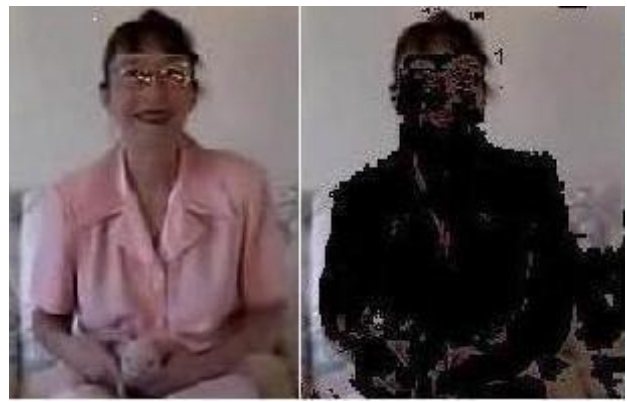

(a) Skin detection example on a frame from Video 12. Black shows skin.

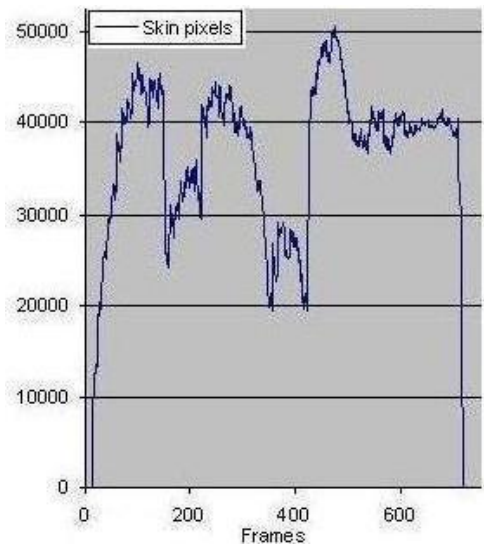

(b) Skin Peaks related to the incorrect skin detection in Video 12.

Fig. 4. Skin detection problems.

PSKIN but incorrectly reported as LSKIN. Figure 4(a) shows clothes and pig detected as skin. The female in Video 12 is wearing pink clothes that match the skin color. Figure 4(b) shows peaks related to the incorrect detection of clothes as skin for Video 12.

Figure 5 is an example frame from Video 24, which is also reported as LSKIN. Figure 5 shows that desert sand is detected as skin. When there is a sufficient match between the color of skin and the color of non-skin objects, the algorithm incorrectly reports it as skin. In such situation, color based skin categorization can be misleading. Texture analysis, use of semantics, and object recognition could help to distinguish skin colored background information from human skin color.

## 5. CONCLUSION

In this paper, we have developed an approach for categorization of videos based on skin color. We have tested our algorithm on 30 test sequences and achieved a true positive rate of over 90 %. In the next step, our goal is acquiring larger collections of videos in order to verify and improve the results.

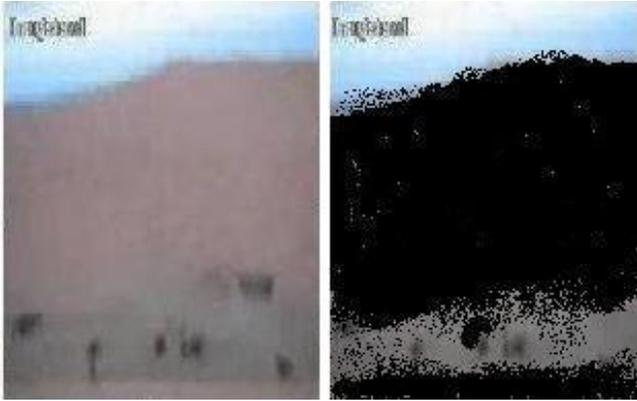

Fig. 5. Skin detection problems. Example frame from Video 24. Sand wrongly detected as skin. Black shows skin.